\documentclass[10pt,twocolumn,letterpaper]{article}

\usepackage{iccv}
\usepackage{times}
\usepackage{epsfig}
\usepackage{graphicx}
\usepackage{amsmath}
\usepackage{amssymb}
\usepackage{algorithm}  
\usepackage{algpseudocode}  
\usepackage{amsmath}  
\usepackage{makecell}
\usepackage{float}
\usepackage{booktabs}
\usepackage{multirow}
\usepackage{subfigure}
\usepackage{soul} 
\usepackage{color, xcolor} 

\usepackage[pagebackref=true,breaklinks=true,letterpaper=true,colorlinks,bookmarks=false]{hyperref}
\usepackage[capitalize]{cleveref}

\iccvfinalcopy 

\def\iccvPaperID{****} 

\ificcvfinal\fi

\begin{document}

\title{Predicting Token Impact Towards Efficient Vision Transformer}

\author{
  Hong Wang\textsuperscript{1} \quad
  Su Yang\textsuperscript{1} \thanks{\textsuperscript{*}Corresponding author: suyang@fudan.edu.cn}\quad
  Xiaoke Huang\textsuperscript{1} \quad
  Weishan Zhang\textsuperscript{2}\\
  \textsuperscript{1} Shanghai Key Laboratory of Intelligent Information Processing, School of Computer Science, Fudan University\\
  \textsuperscript{2} School of Computer Science and Technology, China University of Petroleum\\
  {\tt\small hongwang21@m.fudan.edu.cn}
}
\maketitle
\ificcvfinal\thispagestyle{empty}\fi

\begin{abstract}
   Token filtering to reduce irrelevant tokens prior to self-attention is a straightforward way to enable efficient vision Transformer. This is the first work to view token filtering from a feature selection perspective, where we weigh the importance of a token according to how much it can change the loss once masked. If the loss changes greatly after masking a token of interest, it means that such a token has a significant impact on the final decision and is thus relevant. Otherwise, the token is less important for the final decision, so it can be filtered out. After applying the token filtering module generalized from the whole training data, the token number fed to the self-attention module can be obviously reduced in the inference phase, leading to much fewer computations in all the subsequent self-attention layers. The token filter can be realized using a very simple network, where we utilize multi-layer perceptron. Except for the uniqueness of performing token filtering only once from the very beginning prior to self-attention, the other core feature making our method different from the other token filters lies in the predictability of token impact from a feature selection point of view. The experiments show that the proposed method provides an efficient way to approach a light weighted model after optimized with a backbone by means of fine tune, which is easy to be deployed in comparison with the existing methods based on training from scratch.
\end{abstract}

\section{Introduction}

Transformer as an emerging model for natural language processing \cite{transformer} has attracted much attention in computer vision. So far, a couple of vision Transformers have been proposed and made tremendous success in promising superior performance in a variety of applications compared with convolution neural network based deep learning frameworks \cite{ViT, DeiT, end_to_end_object,bertasius2021space,3DPoint, wang2021pyramid,hudson2021generative, SWin, EnzeXie2021SegFormerSA}. At the same time, a major problem arises: The heavy computational load prevents such models from being applied to edge computing-based applications. Therefore, a recent trend has been shifted to develop light weighted models of vision Transformer (ViT). It is known that self-attention is the major bottleneck to incur dense computations in a Transformer as it requires permutation to couple tokens. Accordingly, the recent efforts were devoted to the following trials: (1) Enforce the self-attention to be confined in a neighborhood around each token such that fewer tokens will be involved in updating each token. The methods falling in this category include Swin Transformer \cite{SWin}, Pale Transformer \cite{Pale}, HaloNet \cite{Halo}, and CSWin Transformer \cite{CSWin}. These methods are based on such an assumption that tokens spatially far away are not semantically correlated, but this does not always hold true. Moreover, since the neighborhood to confine self-attention is predefined, not machine learning based, it may sometimes not be coherent to practice. (2) Another solution aims to modify the self-attention operations internally \cite{LiteViT, rethinkingattention, hydra, ReformerTE}. By changing the computing order in self-attention while incorporating the combination of multiple heads into the self-attention, the complexity of Hydra Attention \cite{hydra} could be made relatively low provided no nonlinear component is contained in the self-attention module, which is a strong constraint to prevent such a solution from being applied broadly. (3) On account of the $O(N^2d)$ complexity of self-attention, where $N$ is the token number and $d$ the feature dimension, a straightforward way is to reduce the number of tokens fed to self-attention instead of the effort to modify self-attention itself. One methodology is to group similar tokens into clusters via unsupervised learning and let each cluster act as a higher-level abstractive representation to take part in the self-attention \cite{cluster1,cluster2}. Here, the difficulty lies in the quality control of clustering, which may lead to not semantically meaningful representations, and thus affect the final decision negatively. The other kind of solution aims to reduce the token number by applying tokens filter explicitly or based on certain heuristics. In \cite{Dynamic}, a couple of token filters realized using multi-layer perceptron (MLP) are incorporated into some middle layers of ViT as gating functions, which are trained end-to-end with the backbone \cite{DeiT,LvViT}, such that the tokens resulting from one self-attention layer can be selectively forwarded to the subsequent self-attention layers. In \cite{AViT}, an early stop criterion based on the accumulated token value at the first dimension is proposed. In \cite{EViT}, token importance is assumed to be its attentive weight correlated to class token. However, the complex coupling layer by layer brings in uncertainty to the attentive weights in terms of correlating to class token, so gradual token filtering has to be applied while the less attentive tokens are also preserved to aid further testing.

In sum, these token filtering methods miss to address the following issue: They are based on heuristics \cite{AViT, EViT} or enclosed in the end-to-end training with backbone \cite{Dynamic}, so the rationality of discarding some tokens selectively is not straightforward. In other words, due to the heuristic and less explainable nature of these methods, they are unable to foresee the impact of a token on the final decision explicitly. Therefore it is impossible for them to filter out all irrelevant tokens from the very beginning and token filtering has to be done gradually in a layer-wise manner, which results in unpredictable token filtering on the fly, not favored by parallel computing.

This study aims to solve the aforementioned problem by proposing a ranking method to measure how relevant a token is in regard to the final decision. Based on such a measure, then, we proceed to train a binary classifier as a token filter with learnable parameters generalized from the whole training corpus, such that we can filter out irrelevant tokens from the very beginning prior to self-attention. For this sake, we propose a measure referred to as delta loss (DL) to evaluate how much the loss changes once masking the token of interest, where the naive Transformer can act as the agent to score the difference of loss caused by with or without a token of interest. The mechanism is similar to a wrapper in the sense of classical feature selection \cite{kohavi1997wrappers}. Then, we label the tokens resulting in big DL values as positive instances since masking them will have a significant impact on the final decision. Further, we train a MLP based binary classifier using the labeled tokens based on their DL values. Finally, we apply such a token filter on each token, prior to all the subsequent Transformer blocks, and fine tune the whole pipeline end-to-end. As a result, the irrelevant tokens can be discarded from the very beginning, which is a one-pass process in contrast to reducing token numbers gradually \cite{Dynamic, AViT, EViT}. 

The contribution of this work is as follows: 

(1) In the context of light weighted ViT, it is the first time that token filtering is proposed from a feature selection point of view to rank the relevance of each token in regard to the final decision. Hence, whether a token makes sense for the final decision becomes predictable from the very beginning, which can prevent irrelevant tokens from taking part in self-attention to the best extent. As a one-pass filter deployed at the very beginning prior to self-attention, it can lead to higher efficiency with even fewer token dropout compared with gradual token dropout throughout the pipeline.

(2) We propose a new metric referred to as delta loss to weigh the importance of each token in terms of affecting the final decision and then force the token classifier to optimize its performance on the pseudo labels quantized from the DL values. 

(3) The only change compared to the original ViT is applying a MLP as the pre-filter for binary classification, which is fine turned with backbone, so the deployment is quite simple compared with the state-of-the-art (SOTA) methods, which rely on training from scratch.

(4) The experiments show that the proposed method promises SOTA performance in terms of both precision and efficiency in an overall sense.

\section{Related works}

\textbf{Vision Transformer. }Transformer is initially applied in natural language processing (NLP) \cite{transformer}. ViT \cite{ViT} is the first work extending Transformer to computer vision by using no-overlapping image patches for image classification such that no convolution operation is needed. It shows comparable performance to convolution neural networks (CNN) on large-scale datasets. To perform well on various vision tasks, however, ViT and some of its following variants \cite{end_to_end_object, 3DPoint, bertasius2021space} require large-scale data and long training time for pre-training. DeiT \cite{DeiT} improved the training configuration by using a novel distillation method and proposed a Transformer architecture that can be trained only with ImageNet1K \cite{imagenet}, whose performance is even better than ViT. 

\textbf{Efficient Transformer.} Although Transformer has recently led to great success in computer vision, it suffers from dense computations arising from self-attention, which is also the major mechanism to grant the promising performance in various down-streaming applications. Therefore, recent efforts are focused on proposing various methods to reduce the self-attention caused by dense computations. Provided there are $N$ tokens of $d$ dimension corresponding with the image patches, the self-attention to correlate every couple from the permutation of the $N$ tokens will result in $O(N^2d)$ complexity in a simple updating round. For deploying Transformer on edge devices, a variety of simplified models have been proposed, aiming to reduce parameters and operations, for example, parameter pruning \cite{channel,prune}, low-rank factorization \cite{rank}, and knowledge distillation \cite{distill1,distill2}. Yet, these strategies for acceleration are limited in that they still rely on CNN, which deviates from the original design of Transformer, that is, facilitating deep learning with a new working mechanism other than CNN. 

One way for rendering light weighted vision Transformer is to simplify the layers of Transformer \cite{zhou2020bert,pan2022less,elbayad2019depth}, but its benefit is limited since the major complexity arises from self-attention, not layer stack. So, some other efforts are focused on altering the internal operations of Transformer to make self-attention more efficient \cite{LiteViT, rethinkingattention, hydra}. As for Hydra Attention \cite{hydra}, the computing order insider self-attention is reorganized while the conbination of multiple heads is incorporated into self-attention to reduce the complexity. Nevertheless, it is workable only when no nonlinear component such as SoftMax is applied in self-attention, which limits its applications.

Some other methods try to alleviate the computations of self-attention by reducing the number of tokens. One way is to enforce the computation of self-attention to be conducted in a predefined local region, for instance, Swin Transformer \cite{SWin}, Pale Transformer \cite{Pale}, HaloNet \cite{Halo}, and CSWin Transformer \cite{CSWin}. These methods are based on the assumption that image patches located far from each other are not semantically relevant, but this only partially holds true. Besides, since determining the local context does not rely on machine learning, it cannot be adaptive to various real scenarios end-to-end. Another solution is grouping similar tokens together to obtain more abstractive sparse token representations from clustering. The self-attention confined to such highly abstractive representations can thus be made efficient. TCFormer \cite{cluster1} fuses the tokens in the same cluster into a new one utilizing a weighted average, and the tokens involved in self-attention can then be reduced layer by layer. When tackling high-resolution images, Liang et al. \cite{cluster2} leverage clustering in the first few layers to reduce the number of tokens and reconstruct them in the last few layers. Thus, the dense computations on self-attention can be avoided in the middle layers. The limit for the clustering based methods is: They simply merge similar tokens but ignore the quality control of token clustering in case some clusters might be spanned by less homogeneous tokens. 

Since the aforementioned approaches suffer from hard quality control or lack of machine learning, this gives rise to another methodology, which aims to filter out tokens gradually throughout the pipeline of ViT. Dynamic ViT \cite{Dynamic} incorporates a couple of learnable neural networks to the middle layers of ViT as the gating structure to make tokens gradually sparser throughout a relatively long course. A-ViT \cite{AViT} calculates the accumulated halting probability of each token by using the feature values resulting from each Transformer layer, which gradually reduces the number of tokens without adding any additional modules, but could result in suddenly halted computing on a token, in general, not favored when scheduling parallel computing. E-ViT \cite{EViT} assumes that top-k attentive weights correspond with relevant tokens but it still preserves irrelevant tokens throughout the whole pipeline to undergo a gradual token dropout procedure. The reason is: Token impact cannot be related to final decision in an explicit way due to the complex inter-layer coupling between tokens when back tracing each token's correlation to class token. Besides, every trail of the hyper parameter k in preserving selectively the top-k attentive tokens will lead to a new-round training from scratch. A common limit of the aforementioned approaches is: All such works rely on the running results of the backbone for token filtering, as it is impossible for them to foresee the token-caused effect on the final decision from the very beginning. In view of such a limit, we propose a new method from a feature selection point of view to conduct token filtering from the very beginning prior to self-attention to filter out truly irrelevant tokens.

\textbf{Feature selection.} In the literature on deep learning, Le et al. \cite{li2016deep} proposed a feature selector by adding a sparse one-to-one linear layer. It directly uses network weight as the feature weight, so it is sensitive to noise. Roy et al. \cite{roy2015feature} used the activation potential as a measure for feature selection at each single input dimension but is limited to specific DNNs. Since then, the interest has been turned to the data with a specific structure, which relies more on the progress of traditional data feature selection methods \cite{grgic2018beyond,liu2016robust}. AFS \cite{gui2019afs} proposes to transform feature weight generation into a mode that can be solved by using an attention mechanism. Takumi et al. \cite{kobayashi2021group} proposed a method that harnesses feature partition in SoftMax loss function for effectively learning the discriminative features. However, these methods are focused on reducing feature map or selecting channels of CNN rather than Transformer. We are the first to use the delta loss value as an indicator for identifying relevant Transformer tokens from a wrapper-based feature selection point of view \cite{kohavi1997wrappers} by testing their impact on the final decision once masked.

\section{DL-ViT}

\begin{figure*}
  \centering
  \includegraphics[width=0.7\textwidth]{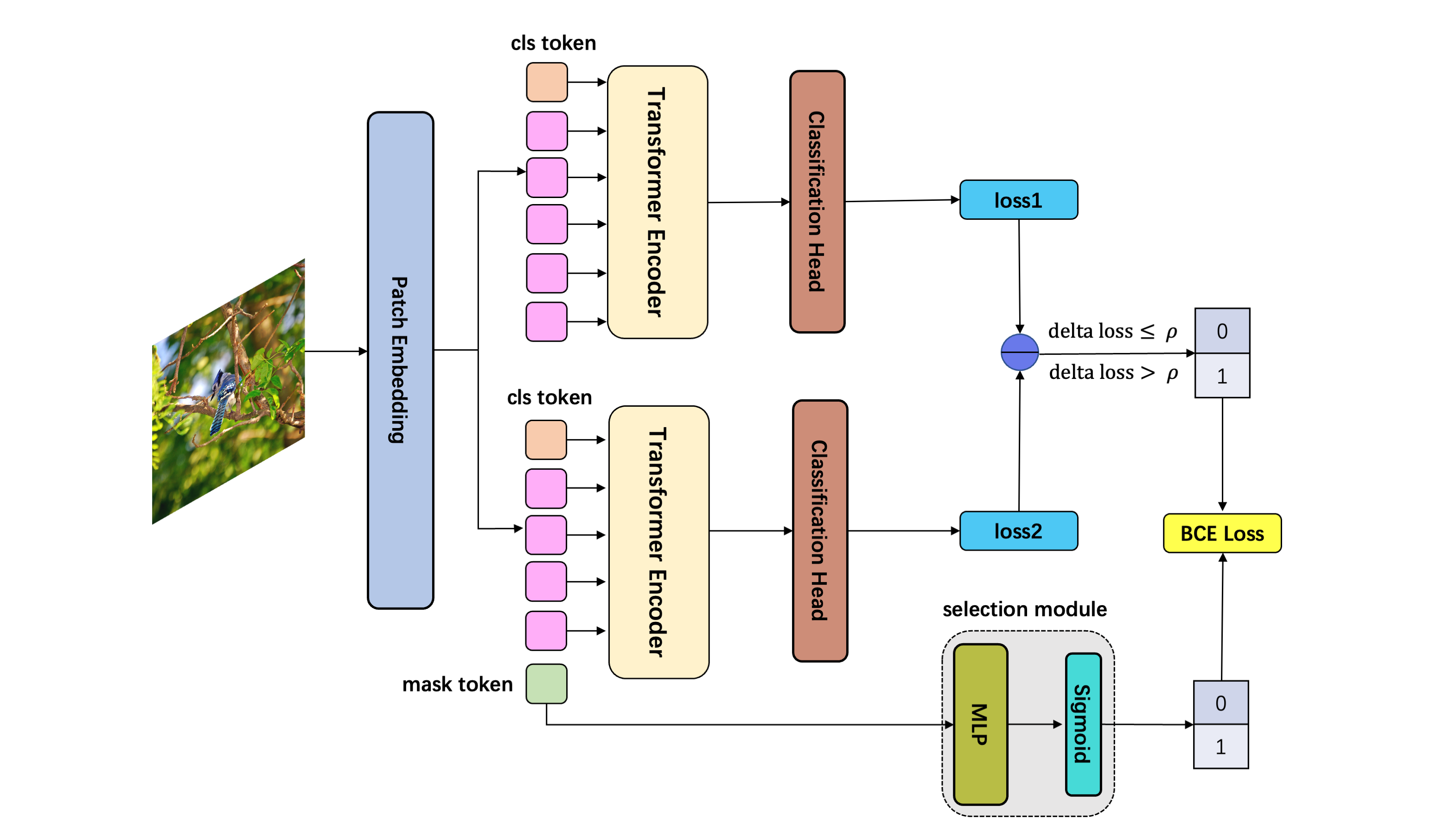}
  \caption{The overall training process: The two branches of the vision Transformer are in fact the same one, whose parameters are fixed during training. The delta loss and $\rho$ refer to \cref{eq:quation7} and \cref{eq:quation8}. }
  \label{fig:figure1}
\end{figure*}

 We propose a metric referred to as delta loss to weigh how vital a token is. In detail, we mask a token at first and then compute its impact on the loss, say, the change of cross entropy with/without such token for the final decision. If masking a token leads to big DL, it means that such a token does affect the final decision much, which should be preserved to take part in the subsequent self-attentions. Vice versa, if the loss does not change much with/without a token, such a token should be discarded due to its less importance to the decision. Correspondingly, a plausible trick arising from the aforementioned scheme is: The Transformer itself can act as the agent to score the importance of each token via DL without any further machine learning required in this phase. By using the DL scores to label the tokens in the training corpus as positive or negative, we can then train a binary classifier to check whether the tokens of an input image should be preserved to take part in the subsequent self-attentions, where the classifier is implemented using MLP. Finally, we preset the simple MLP module prior to the backbone Transformer, and fine tune the whole pipeline, where preset a token filter as such is the only change in the architecture. 

In the following, we describe the two phases of the delta loss based efficient vision Transformer (DL-ViT): Evaluating token importance with delta loss to train the token filter and then fine tuning the entire network after incorporating the token filter. After an image passes through the embedding layer of the vision Transformer, the non-overlapping image patches are encoded into tokens denoted as:
\begin{equation}
  \vspace{1pt}
  X=\left\{x_i\in \mathcal{R} ^d|i=1,2…,N\right\},
  \label{eq:quation1}
\end{equation}
where $N$ is the total number of tokens and $d$ the embedding dimension. After masking the $i$-th token $x_i\in \mathcal{R}^d$, we get the tokens in the following form:
\begin{equation}
  X_i=\left\{x_1,…x_{i-1},\varnothing,x_{i+1},…,x_N\right\},
  \label{eq:quation2}
\end{equation}
where $\varnothing$ means replacing the $i$-th token with zeros (masking). Then, we feed $X$ and $X_i$  to the Transformer, respectively, to obtain the corresponding prediction results:
\begin{equation}
 \vspace{1pt}
 \hat{y} = Transformer(X),
  \label{eq:quation3}
\end{equation}
\begin{equation}
  \hat{y}_i = Transformer(X_i).
  \label{eq:quation4}
\end{equation}
Based on the previous prediction results, we calculate the cross-entropy loss of either case in reference to the ground truth y as follows:
\begin{equation}
 \mathcal{L} = CrossEntropy(\hat{y}, y),
  \label{eq:quation5}
\end{equation}
\begin{equation}
 \vspace{1pt}
  \mathcal{L}_i = CrossEntropy(\hat{y}_i, y).
  \label{eq:quation6}
\end{equation}
It is known that the value of loss measures how close the prediction result approaches the ground truth, where a lower value corresponds with closer to the ground truth. Let:
\begin{equation}
  \Delta\mathcal{L}_i = \mathcal{L}-\mathcal{L}_i.
  \label{eq:quation7}
\end{equation}
Obviously, if the delta loss defined in \cref{eq:quation7} is positive, it means that masking the $i$-th token makes the decision closer to the ground truth since masking as such causes a lower cross-entropy value in contrast to the original case. In such a case, discarding the token should not affect but benefit the decision of ViT, and a bigger delta loss corresponds with a better change on the decision. So, we quantize the delta loss measure to mark whether the current token should be discarded or not, formulated as:
\begin{equation}
  label(x_i)=
  \begin{cases} 
    0,  & \mathcal{L}-\mathcal{L}_i \leq \rho\\
    1, & \mathcal{L}-\mathcal{L}_i > \rho
  \end{cases}
  \label{eq:quation8}
\end{equation}
where 0 means leaving the token out, 1 preserving the token, and $\rho$ the only hyperparameter to control the significance of the pseudo labeling.

After labeling all the tokens in the training corpus, we can then proceed to learn the generalizable law to distinguish positive token examples from negative ones in a population sense, which leads to a binary classifier realized using MLP for token filtering, acting to determine whether each token should be preserved to the next phase of the pipeline or not. So far, there is still a critical problem to be tackled, that is, some similar tokens may lead to contradicting results in terms of delta loss. This is quite common when two images share some similar patches locally but are quite different in an overall sense. Such semantically ambiguous local patches impose difficulty on token filter training, so we attach the profile featuring the whole image to each token as context, namely, global feature, to solve this problem. That is, we not only use the tokens with original embedding but also apply adaptive average pooling (AAP) over all tokens of an image to obtain the global feature of the image, acting as the context to make each token distinguishable from the others. Thus, the overall descriptor for each token becomes:
\begin{equation}
  x_i^\prime=[x_i, x_{global}].
  \label{eq:quation9}
\end{equation}
\begin{equation}
  x_{global} = AAP(X) = \frac{1}{N}\Sigma^N_{k=1}x_k.
  \label{eq:quation10}
\end{equation}
Consequently, $x_i^\prime$ instead of $x$ is fed to the token selection module for training and inference:
\begin{equation}
  p_i = Sigmoid(MLP(x_i^\prime)).
  \label{eq:quation11}
\end{equation}

During training, we first fix all parameters of the pre-trained backbone Transformer for token labeling, and then, train the MLP only. Here, we use binary-cross-entropy loss to train the network:

\begin{equation}
 \mathcal{L}_{MLP} = BinaryCrossEntropy(p_i, label(x_i)),
  \label{eq:quation12}
\end{equation}

where $p_i$ is the prediction from MLP, and $label(x_i)$ the pseudo label calculated from \cref{eq:quation8}. \cref{fig:figure1} depicts how to train the token selection module with delta loss. \cref{alg:label} and \cref{alg:train} describe respectively how to label tokens with naive DeiT \cite{DeiT} and how to train the selection module.

  \begin{algorithm}[htb]
  \caption{ Token labeling with naive DeiT \cite{DeiT}}
  \label{alg:label}
  \begin{algorithmic}[1]
    \Require
      $\textbf{X}=\left\{x_i\in \mathcal{R} ^d|i=1,2…,N\right\}$,
      and the corresponding ground truth $y$.
    \Ensure
      $\textbf{Label}=\left\{label(x_i) |i=1,2…,N\right\}$.
    \State $\textbf{Label} = \varnothing$
    \State Set $\rho$ to control the significance of pseudo labeling.
    \State $\hat{y} = Transformer(X)$
    \State $\mathcal{L} = CrossEntropy(\hat{y}, y)$
    \For  {$i = 1, 2, ..., N$}
    \State $X_i=\left\{x_1,…x_{i-1},\varnothing,x_{i+1},…,x_N\right\}$
    \State $\hat{y}_i = Transformer(X_i)$
    \State $\mathcal{L}_i = CrossEntropy(\hat{y}_i, y)$
    \If  {$\mathcal{L}-\mathcal{L}_i \leq \rho$}
    \State $label(x_i)=0$
    \Else
    \State $label(x_i)=1$
    \EndIf
    \State $\textbf{Label} =\textbf{ Label} \cup label(x_i)$
    \EndFor
    \\
    \Return $\textbf{Label}$
  \end{algorithmic}
\end{algorithm}

  \begin{algorithm}[htb]
  \caption{ Token filter training}
  \label{alg:train}
  \begin{algorithmic}[1]
    \Require
      Batch of images with tokens and the corresponding pseudo labels in the form of $\textbf{X}=\left\{x_i\in \mathcal{R} ^d|i=1,2…,N\right\}$
      and $\textbf{Label}=\left\{label(x_i) |i=1,2…,N\right\}$.
    \Ensure
      Parameters $\textbf{W}$ of the MLP token filter.
    \State Random initialization of \textbf{W}
    \Repeat
    \State Load \textbf{(X,Label)} of one image in the batch in turn
    \State $x_{global}=AAP(X)$
    \For{i = 1, ..., N}
    \State $x_i^{\prime}=[x_i,x_{global}]$
    \State $p_i = Sigmoid(MLP(x_i^\prime))$
    \State $\mathcal{L}_{MLP} = BinaryCrossEntropy(p_i, label(x_i))$
    \State Back-propagation updating \textbf{W}
    \EndFor
    \Until{no more descent on $L_{MLP}$}
    \\
    \Return $\textbf{W}$
  \end{algorithmic}
\end{algorithm}
\begin{figure}
  \centering
  \includegraphics[width=0.48\textwidth]{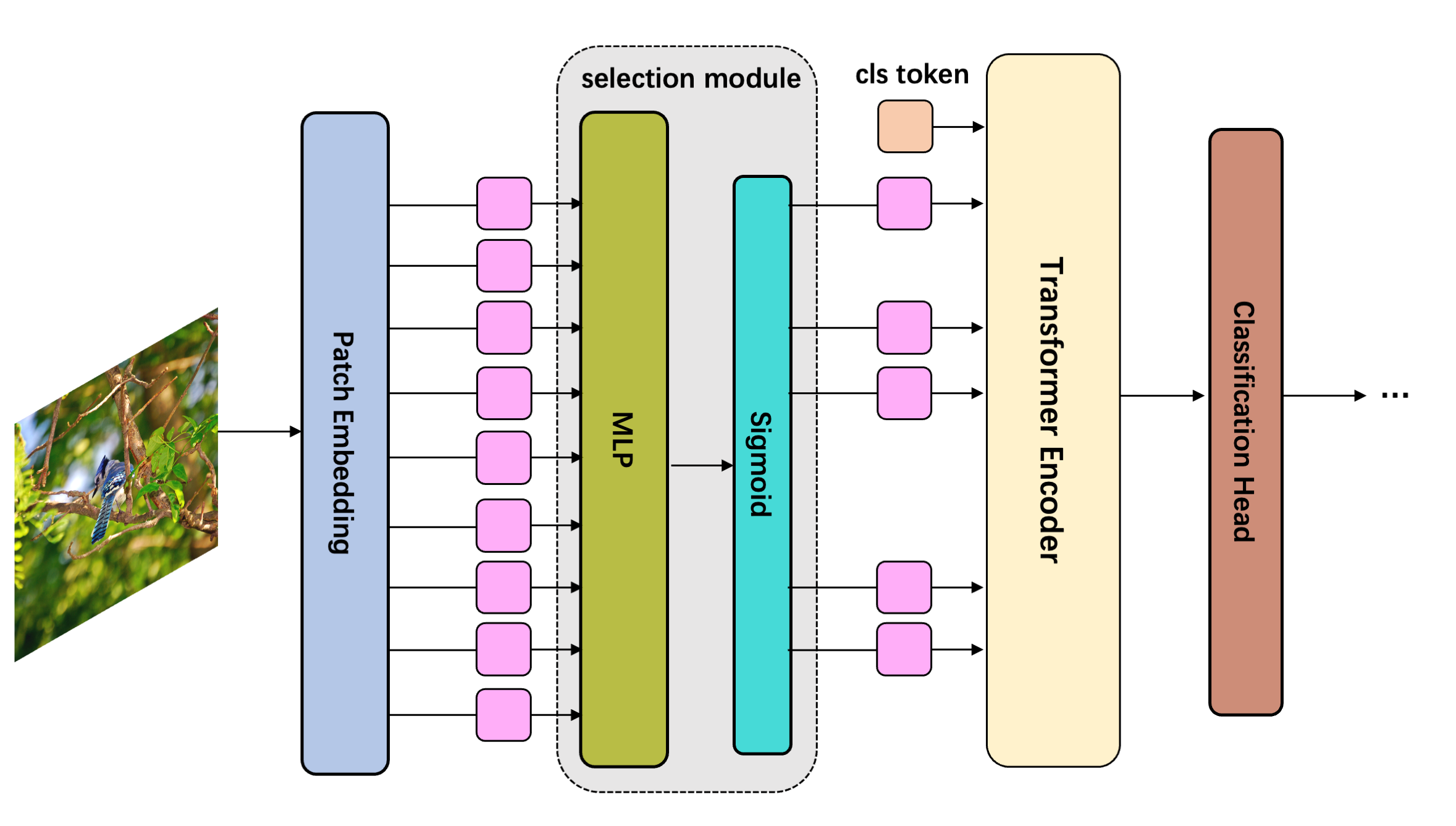}
  \caption{The overall fine tuning and inference process of the proposed approach. All the tokens enter the selection module in turn to decide whether they should be passed to the subsequent pipeline of Transformer according to the predicted probability, after which the number of the preserve tokens will remain unchanged in the rest pipeline.}
  \label{fig:figure2}
\end{figure}

As shown in \cref{fig:figure2}, before entering the Transformer, all tokens must go through the token selection module that will output the decision of keeping or discarding the token. During fine tuning, we train both the token selection module and the DeiT end-to-end, based on the cross-entropy loss:
\begin{equation}
 \mathcal{L}_{finetune} = CrossEntropy(\hat{y}, y),
  \label{eq:quation13}
\end{equation}
where $y$ is the ground truth and $\hat{y}$ the output of the whole network. 

\begin{figure*}[htbp]
	\centering
	\begin{minipage}{0.45\linewidth}
		\centering
		\includegraphics[width=0.55\linewidth]{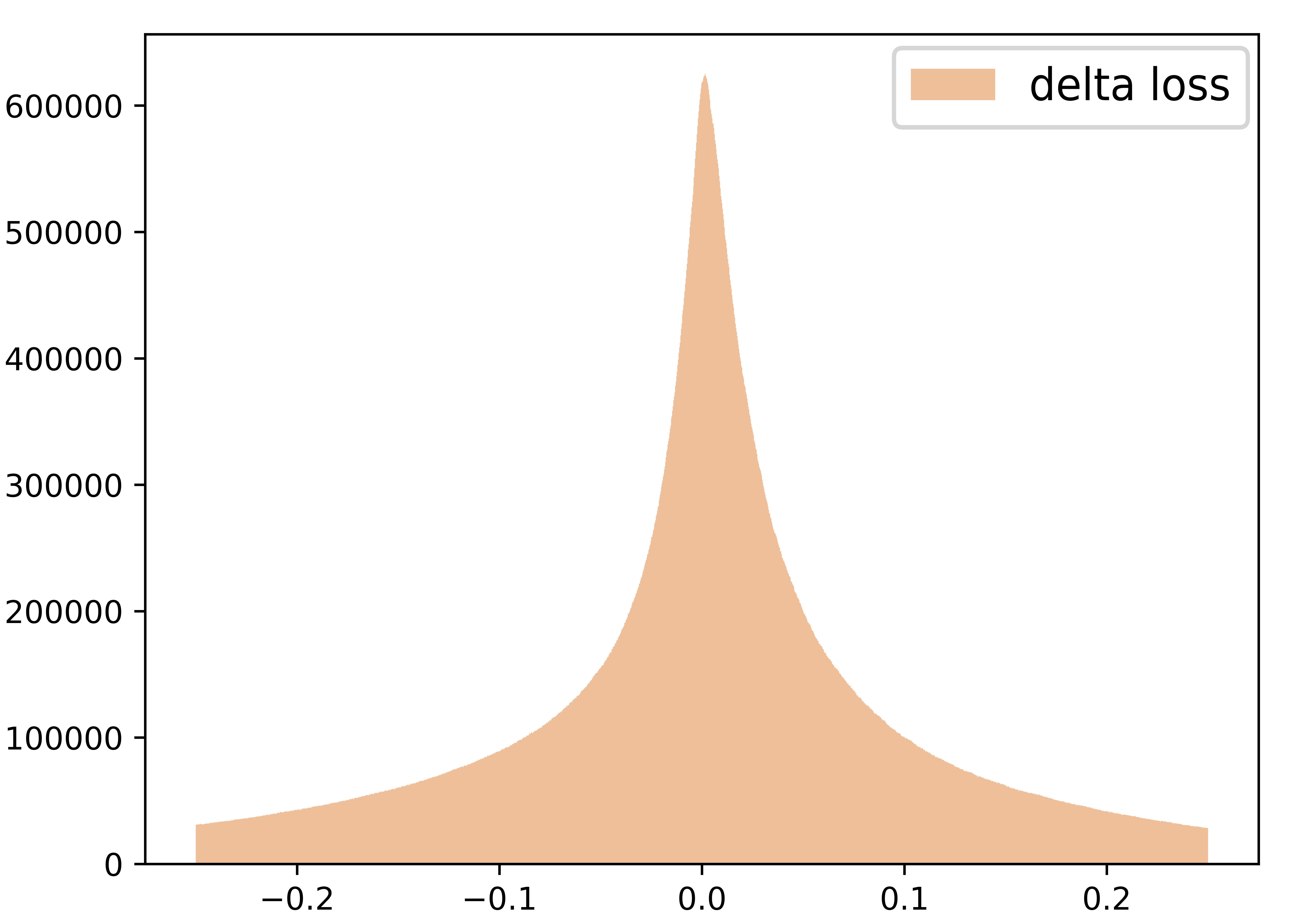}
		\caption{Distribution of all the DL values on ImageNet1K training set.}
		\label{fig:loss-b}
	\end{minipage}
	\begin{minipage}{0.45\linewidth}
		\centering
		\includegraphics[width=0.55\linewidth]{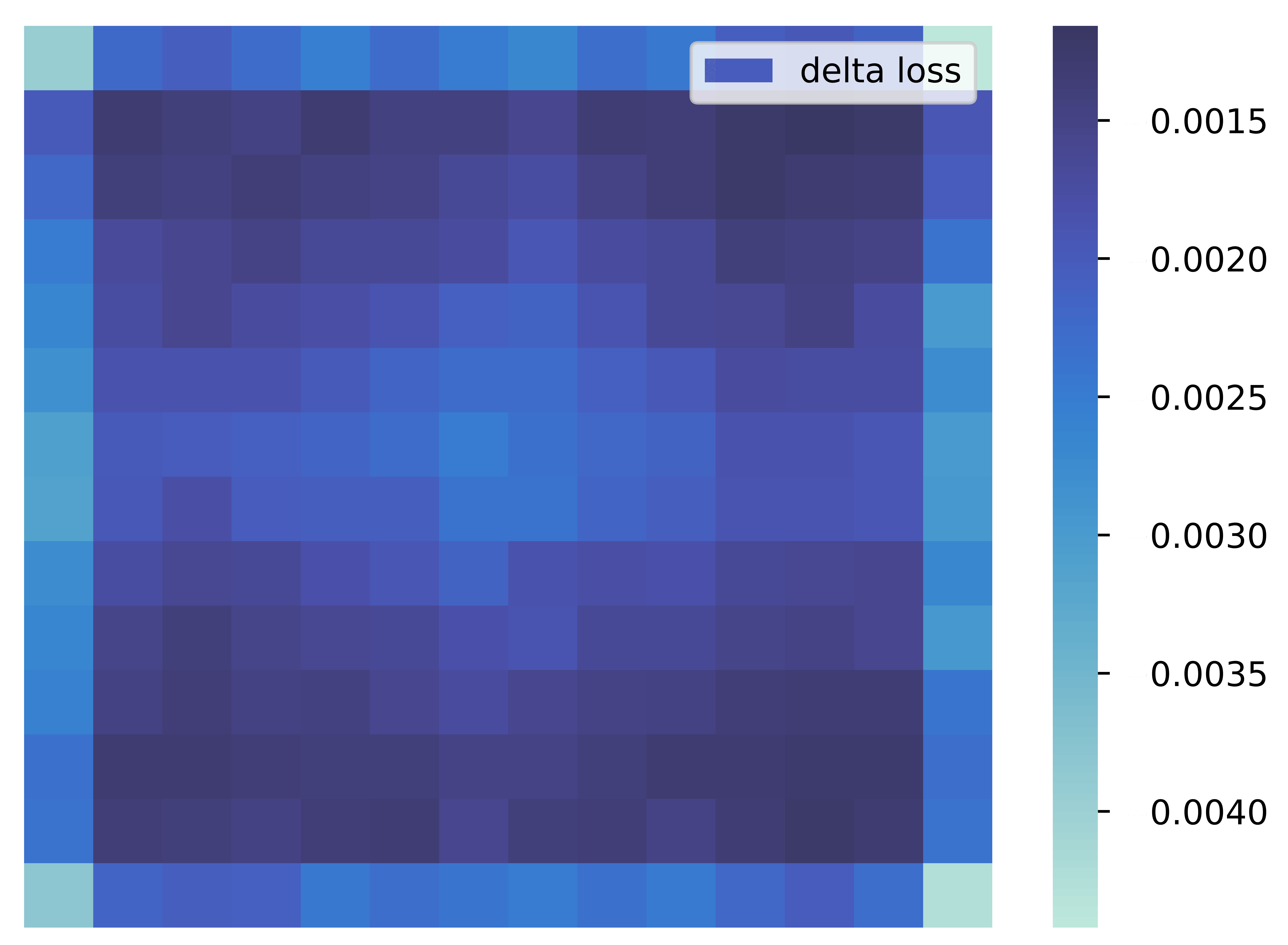}
		\caption{Average DL of every image patch obtained from DeiT-T \cite{DeiT}. Darker color corresponds with lower DL value.}
		\label{fig:loss-a}
	\end{minipage}
\end{figure*}

During fine tuning, in order to make it easy to parallelize the computation, we do not delete tokens directly but replace them with zeros to prevent them from affecting subsequent operations. Such a token masking strategy makes the computational cost of the training iterations similar to those of the original vision Transformer. During inferring, we throw the masked tokens out of the subsequent calculations in order to examine the actual acceleration resulting from the token selection mechanism.

\section{Experiments}
\textbf{Data:} We evaluate our method for image classification on the 1000-class ImageNet1K ILSVRC 2012 dataset \cite{imagenet}, and all images have a resolution of $224 \times 224$. 

\textbf{Experimental setting:} Following the baselines \cite{AViT, Dynamic,EViT}, we use the data-efficient vision Transformer (DeiT) \cite{DeiT} as the backbone, and following its training principles, we only use the ImageNet1K dataset for training. We use $16 \times 16$ patch resolution and SGD optimization. The MLP is composed of $3$ layers with ReLU for the first two layers and Sigmoid for the last layer as the activation and the number of neurons are set to $384$, $100$, and $1$ for each layer, respectively. When training MLP, we use the pre-trained model of DeiT to compute the loss value, and the learning rate is fixed to $1\times10^{-2}$. When fine tuning the whole network, the learning rate is $1\times10^{-3}$ and reduced by 10 times for every 40 epochs. For regularization, we set the weight decay of the optimizer to $1\times10^{-4}$ in both MLP training and fine tuning. Starting from publicly available pre-trained checkpoints and the pre-trained token selection module, we fine tune the DL-ViT-T/S variant models evolved from DeiT-T/S \cite{DeiT} for 100 epochs, where T/S refers to 3-head/6-head with 192-dimension/384-dimension implementation on 12 layers, respectively. We use 2 NVIDIA 3090 GPUs for training.

\subsection{Intuitive insight from statistics}
\begin{figure}[ht]
  \centering
  \includegraphics[width=0.5\linewidth]{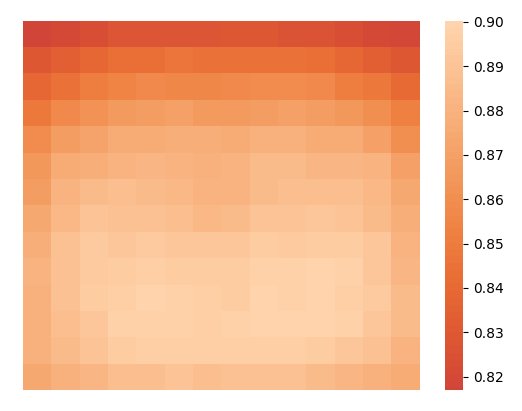}
   \caption{The average of masks predicted by our token selection module on ImageNet1K validation set.}
   \label{fig:mask}
\end{figure}

\begin{figure*}[ht]
    \centering
    \vspace{3pt}
    \includegraphics[width=0.65\textwidth]{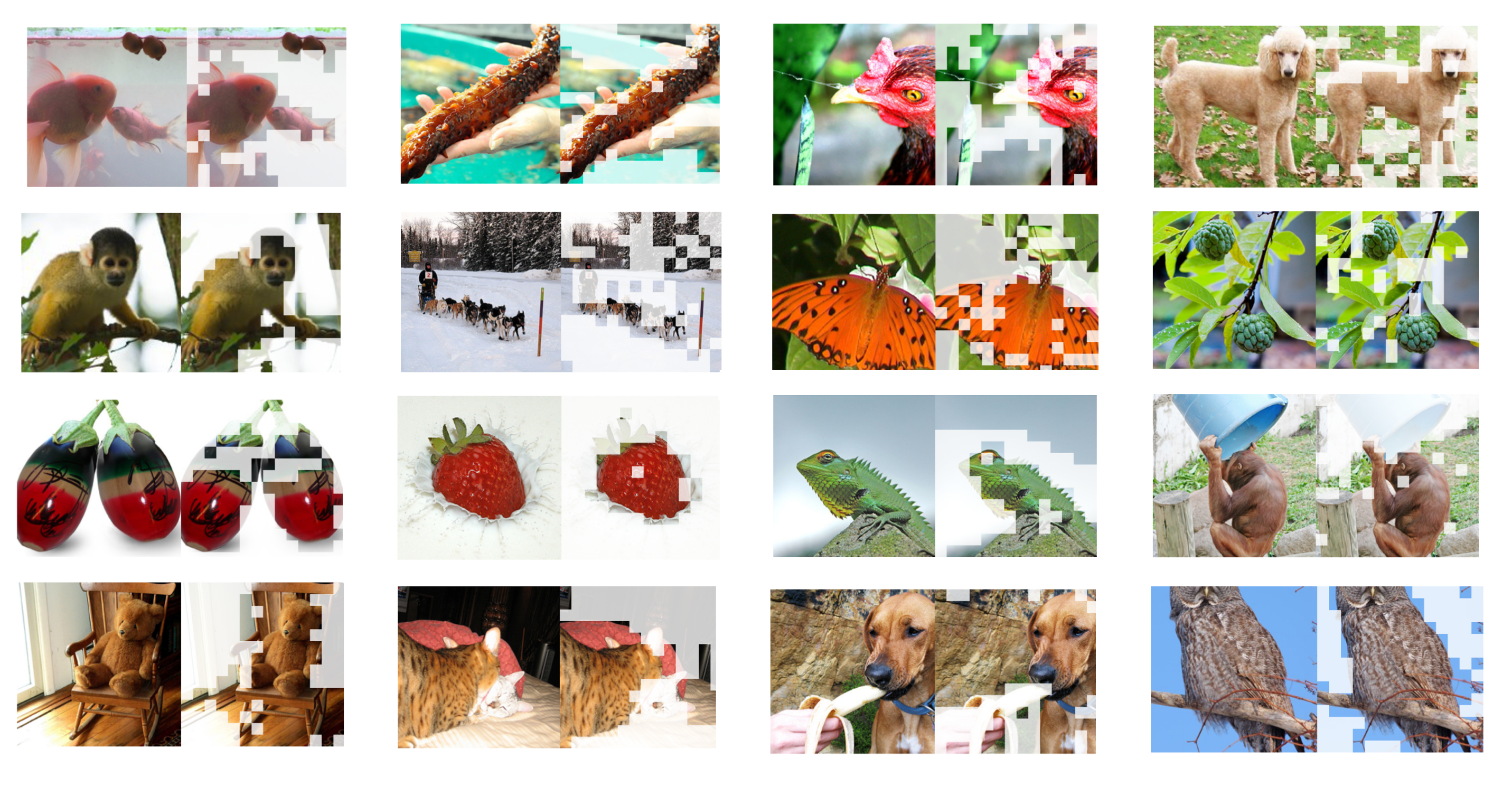}
    \caption{Original image (left) and the masked image (right) resulting from DL-ViT on the ImageNet1K set. The left two columns are the validation set, and the right two columns the training set.}
    \label{fig:show}
\end{figure*}

\begin{figure*}[htbp]
	\centering
	\begin{minipage}{0.45\linewidth}
		\centering
		\includegraphics[width=0.85\linewidth]{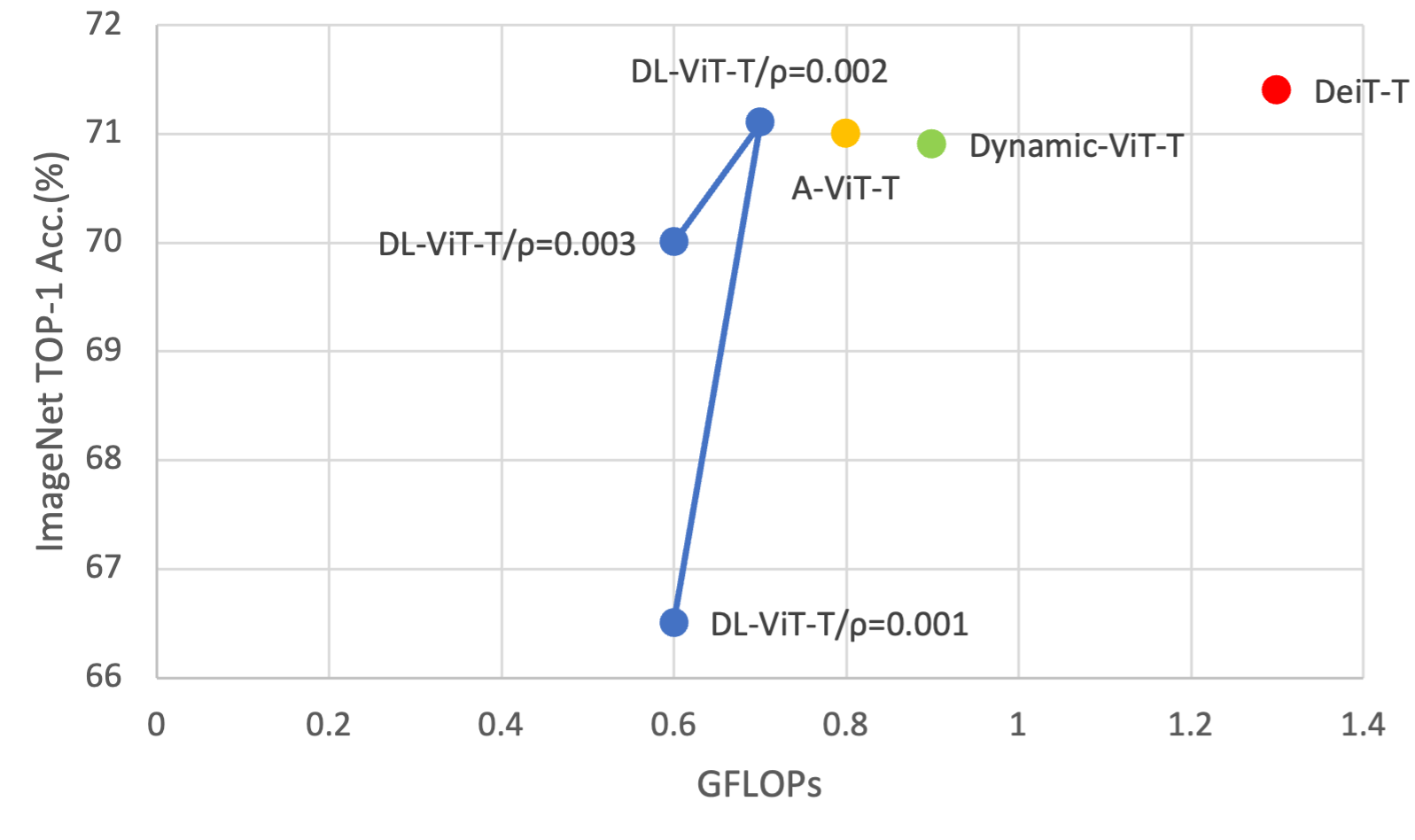}
		\caption{The tiny model complexity (FLOPs) and top-1 accuracy trade-offs on ImageNet.}
		\label{fig:xian-a}
	\end{minipage}
	\begin{minipage}{0.45\linewidth}
		\centering
		\includegraphics[width=0.85\linewidth]{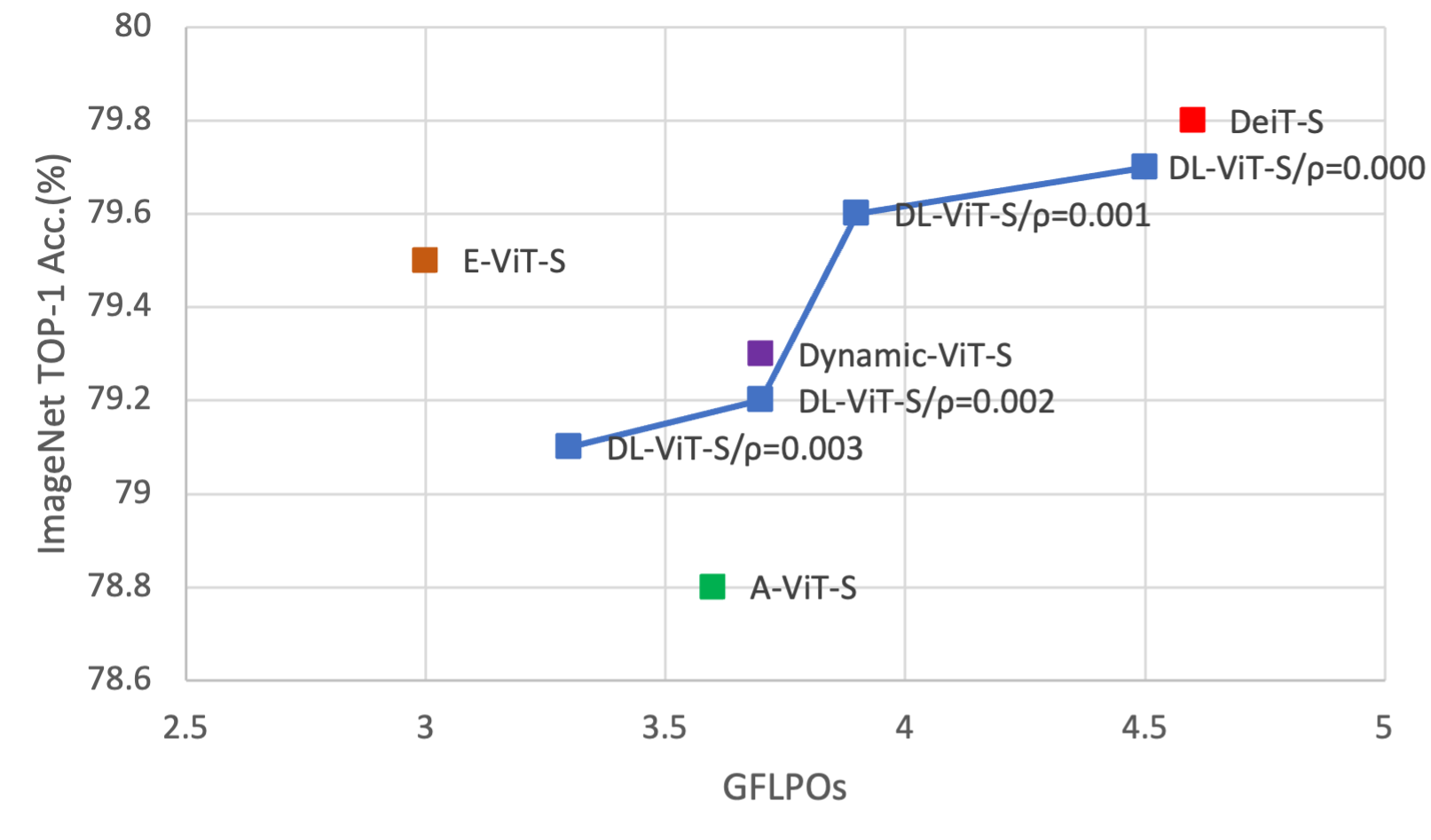}
		\caption{The small model complexity (FLOPs) and top-1 accuracy trade-offs on ImageNet.}
		\label{fig:xian-b}
	\end{minipage}
\end{figure*}

\textbf{Distribution of DL values:} In order to examine the rationality of our method intuitively, we visualize the distribution of all DL values in \cref{fig:loss-b}. We find that small DL values around $0$ dominate the majority of the distribution, which reveals the fact that only a part of the image patches are significantly relevant to classification. So considerable tokens with smaller DL values can be discarded.

\cref{fig:loss-a} depicts the average DL value of the tokens at each patch resulting from DeiT-T \cite{DeiT} on the training set of ImageNet1K. We find that most semantically important patches are on the center of an image, the rest of which should fall into the outliers to be eliminated more frequently by our method.

\begin{table*}[ht]
  \centering
  \begin{tabular}{cccccc}
    \toprule
    \multirow{2}*{Model} & \multicolumn{3}{c}{Efficiency} &\multirow{2}*{Top1 Acc.(\%) $\uparrow$} & \multirow{2}*{Resolution} \\
    \cmidrule{2-4} 
    & \#Params.$\downarrow$ & FLOPs $\downarrow$ & Throughput $\uparrow$ \\
    \midrule
    ViT-B \cite{ViT} & 86.0M & 17.6G& 563 imgs/s& 77.9 & 224\\
    \midrule
    DeiT-S \cite{DeiT} & 22.0M & 4.6G& 1500 imgs/s& 79.8 & 224\\
    Dynamic-ViT-S \cite{Dynamic} & 22.7M & 3.7G & 1654 imgs/s&79.3 & 384\\
    A-ViT-S \cite{AViT} & \textbf{22.0M} & 3.6G & 1849 imgs/s& 78.8 & 224\\
    E-ViT-S \cite{EViT} & 22.1M&\textbf{3.0G}& \textbf{1923 imgs/s}& 79.5&224\\
    DL-ViT-S(ours) & 22.1M& 3.9G& 1602 imgs/s& \textbf{79.6}&224\\
    \midrule
    DeiT-T \cite{DeiT} & 5.7M &1.3G& 3231 imgs/s&71.4 & 224\\
    Dynamic-ViT-T \cite{Dynamic} & 5.9M & 0.9G & 4361 imgs/s& 70.9 & 224\\
    A-ViT \cite{AViT} & 5.7M & 0.8G & 4523 imgs/s & 71.0 & 224\\
    DL-ViT-T(ours) & \textbf{5.7M} & \textbf{0.7G} & \textbf{4565 imgs/s}&\textbf{71.1} & 224\\
    \bottomrule
  \end{tabular}
  \caption{Comparison with baselines. Except for E-ViT, which undergoes training of 300 epochs, the other models are trained with 100 epochs. Note that Dynamic-ViT-S turns out from the resolution of 384 $\times$ 384.}
  \label{tab:table1}
\end{table*}

\textbf{Qualitative analysis.} \cref{fig:mask} visualizes the masks on the ImageNet1K validation set resulting from DL-ViT, where the dark portion appearing mostly along the edge of an image are the less contributive patches for classification, namely, the outliers favored by the token filter to activate elimination. Sometimes, the token selection module eliminates not only the background of the image, but also the confusing portion that may cause classification errors. For example, the third image in the last row of \cref{fig:show} prefers eliminating the patches unrelated to the dog.

\subsection{Comparison to baselines}
We compare our method with the baselines in \cref{tab:table1} in terms of efficiency and precision, where we set $\rho$ to 0.002 and 0.001 for DL-ViT-T and DL-ViT-S, respectively. At the cost of sacrificing only $0.3\%$ and $0.2\%$ accuracy compared with that of the backbone, we cut down $46\%$ and $15\%$ FLOPs of DeiT-T and DeiT-S, respectively. Moreover our method performs best to make DeiT-T more efficient and more precise compared with the baselines, where the Floating-point Operations (FLOPs) metric is measured by FlopCountAnalysis\cite{github}.

\begin{table*}[ht]
  \centering
  \begin{tabular}{ccccccccc}
    \toprule
    Threshold &  \makecell[c]{Top-1 Acc.\\(\%)$\uparrow$} & \makecell[c]{Top-5 Acc.\\(\%)$\uparrow$} & FLOPs $\downarrow$& \makecell[c]{Throughput\\(images/s)$\uparrow$} &  \makecell[c]{Top-1 Acc.\\(\%)$\uparrow$} & \makecell[c]{Top-5 Acc.\\(\%)$\uparrow$} & FLOPs $\downarrow$ & \makecell[c]{Throughput\\(images/s)$\uparrow$}\\
    \midrule
    DeiT-T \cite{DeiT} & 71.4 & 90.8& 1.3G & 3231&71.4& 90.8&1.3G & 3231\\
    \midrule
    & \multicolumn{4}{c}{\makecell[c]{DL-ViT-T with local feature}} & \multicolumn{4}{c}{\makecell[c]{DL-ViT-T with local and global feature}} \\
    \cmidrule{2-9} 
    0.001 &   72.0  & 90.8  &   0.8G&  4771& 66.5& 86.1& 0.6G&4690\\
    0.002 & \textbf{73.1} & \textbf{91.5} & 1.0G& 3937& \textbf{71.1} & \textbf{89.9} & 0.7G & 4565\\
    0.003 & 62.4 & 83.5 & \textbf{0.3G}&  \textbf{5996}& 70.0 & 89.1 & \textbf{0.6G} & \textbf{5527}\\
    \bottomrule
  \end{tabular}
  \caption{Performance of DL-ViT-T subject to local/global feature and threshold $\rho$.}
  \label{tab:table2}
\end{table*}

As E-ViT is an exception that only reports comparison on ViT-S, we follow A-ViT \cite{AViT} and Dynamic-ViT \cite{Dynamic} to report the performance on both ViT-S and ViT-T. Regarding ViT-S, no method performs best on all metrics, where E-ViT runs faster but is inferior to DL-ViT on top-1 precision. Except for the highest top-1 precision on both benchmarks, DL-ViT promises the state-of-the-art (SOTA) performance in an overall sense if taking into account both benchmarks. Since E-ViT misses to compare with all the baselines on ViT-T, except for the performance, we compare it with DL-ViT in a methodological sense to allow a more comprehensive insight: (1) We evaluate token importance via delta loss while E-ViT leverages top-k attention weights as token importance; (2) We filter out irrelevant tokens from the very beginning but E-ViT does this gradually and preserve both important and less important tokens in the whole pipeline. That is, E-ViT cannot foresee the impact of each token on the final decision at the beginning but DL-ViT can. (3) E-ViT modifies the self-attention, and the whole pipeline has to be changed wherever attention is applied, so it has to train from scratch, which is too expensive compared with the fine tune as adopted in our framework. As we change nothing in ViT, the pseudo labeling is performed by using naive ViT without any training. Besides, MLP is a two-class classifier, whose training is not tough. In this sense, the change on the architecture is minor. (4) In DL-ViT, $\rho$ controls the significance of pseudo labeling, where the heuristics to choose its value lies in the statistics of DL values as shown in \cref{fig:loss-b}. For E-ViT, determining k is not easy in that there is no explicit heuristic to foresee its impact on the overall performance, and every trail will lead to a new-round computation-intensive training from scratch. Besides, the layer-varying token importance accounts for why layer-wise token dropout has to be done gradually.
In \cref{fig:xian-a} and \cref{fig:xian-b}, we compare DL-ViT with the baselines under different settings. It is obvious that our model can achieve a good trade-off between efficiency and precision.

In addition to FLOPs, we also evaluate the image throughput of our model on a single NVIDIA RTX 3090 GPU with batch size fixed to 64, and for GPU warming up, 512 forward passes are conducted. The experiment demonstrates that our DL-ViT can accelerate the inference by $15\%\sim 41\%$.

\subsection{Ablation study}

\begin{table}[ht]
  \centering
  \begin{tabular}{cccc}
    \toprule
    \multirow{2}*{Strategy} & \multicolumn{3}{c}{Metric} \\
    \cmidrule{2-4}
        & \#Params.$\downarrow$ & FLOPs $\downarrow$ & Top1 Acc.(\%) $\uparrow$\\
    \midrule
    DL-ViT-T & \textbf{5.7M} & 0.7G& \textbf{71.1}\\
    \midrule
    \makecell[c]{DeiT-T$_0$ \cite{DeiT}} & 5.7M& 0.7G& 68.2 \\
    \midrule
    \makecell[c]{DL-ViT-T$_0$} & 5.7M&  \textbf{0.4G}& 56.6\\
    \bottomrule
  \end{tabular}
  \caption{Comparison with DeiT using random token discard (DeiT-T$_0$ in the second row) and DL-ViT without pre-training but randomly initializing the MLP (DL-ViT-T$_0$ in the third row). }
  \label{tab:table3}
\end{table}

In our method, the full configuration of a solution is subject to the following factors: The backbone for token importance evaluation, the threshold $\rho$ to control the annotation on DL values, MLP, and the local/global feature applied to it. As shown in \cref{tab:table2}, a high threshold value of $\rho$ can filter out more tokens, resulting in higher efficiency, but a too high one will cause degradation in precision. So, there is a compromise to determine the value of $\rho$, where we let $\rho = 0.002$ for DL-ViT-T. Note that when $\rho=0.002$, the accuracy of using only local features is even higher than that of DeiT-T \cite{DeiT}, at the cost of sacrificing FLOPs. Yet, we incorporate global feature as our primary solution due to its promising overall performance. Note that both cases lead to varying accuracy when $\rho$ changes from 0.001 to 0.003, but the FLOPs with global feature remain stably low.

\cref{tab:table3} shows that the proposed model degrades in terms of precision if replacing the pre-training of MLP with the random initialization, and the backbone based on random token filtering also leads to inferior performance. This indicates that our token filtering scheme does contribute to making the DeiT-T efficient while preserving its precision to the best extent.
\section{Conclusions}
We develop an efficient vision transformer with token impact prediction such that token filtering can be deployed at the very beginning prior to self-attention, where the backbone Transformer is used as an agent/wrapper to rank the impact in terms of the difference of loss caused by masking a token of interest. It is the first time to develop a light-weighted model from a feature selection point of view with explicit insight into token’s relevance to the decision. A MLP for token filtering is the only added module, which acts as a two-class classifier with minor change on the overall architecture, and its training is not tough. The present solution is a one-pass filter. In the future, we will investigate into the relevance of tokens at middle layers to the final decision to further improve the efficiency.

{\small
\bibliographystyle{ieee_fullname}
\bibliography{egbib}
}

\end{document}